\documentclass[sn-mathphys,Numbered]{sn-jnl}

\usepackage{graphicx}%
\usepackage{multirow}%
\usepackage{amsmath,amssymb,amsfonts}%
\usepackage{mathtools}      
\usepackage{amsthm}%
\usepackage{mathrsfs}%
\usepackage[title]{appendix}%
\usepackage{xcolor}%
\usepackage{textcomp}%
\usepackage{manyfoot}%
\usepackage{booktabs}%
\usepackage{algorithm}%
\usepackage{algorithmicx}%
\usepackage{afterpage}
\usepackage{algpseudocode}%
\usepackage{listings}%
\usepackage{tabularx}
\usepackage{bookmark}
\usepackage{longtable}
\usepackage{anyfontsize}
\usepackage{xcolor}
\usepackage{siunitx}
\usepackage{subcaption}

\begin{document}

\title[Predicting Change, Not States: An Alternate Framework for Neural PDE Surrogates]{Predicting Change, Not States: An Alternate Framework for Neural PDE Surrogates}

\author*[1]{\fnm{Anthony} \sur{Zhou}}\email{ayz2@andrew.cmu.edu}

\author*[1,2]{\fnm{Amir} \sur{Barati Farimani}}\email{barati@cmu.edu}

\affil[1]{\orgdiv{Department of Mechanical Engineering}, \orgname{Carnegie Mellon University}, \orgaddress{\city{Pittsburgh}, \state{PA}, \country{USA}}}
\affil[2]{\orgdiv{Machine Learning Department}, \orgname{Carnegie Mellon University}, \orgaddress{\city{Pittsburgh}, \state{PA}, \country{USA}}}

\abstract{Neural surrogates for partial differential equations (PDEs) have become popular due to their potential to quickly simulate physics. With a few exceptions, neural surrogates generally treat the forward evolution of time-dependent PDEs as a black box by directly predicting the next state. While this is a natural and easy framework for applying neural surrogates, it can be an over-simplified and rigid framework for predicting physics. In this work, we evaluate an alternate framework in which neural solvers predict the temporal derivative and an ODE integrator forwards the solution in time, which has little overhead and is broadly applicable across model architectures and PDEs. We find that by simply changing the training target and introducing numerical integration during inference, neural surrogates can gain accuracy and stability in finely-discretized regimes. Predicting temporal derivatives also allows models to not be constrained to a specific temporal discretization, allowing for flexible time-stepping during inference or training on higher-resolution PDE data. Lastly, we investigate why this framework can be beneficial and in what situations does it work well. }


\maketitle

\section{Introduction}\label{sec:introduction}
Partial differential equations (PDEs) model a wide variety of physical phenomena, from neuron excitations to fluid flows to climate patterns. Many of the most complex yet practically relevant PDEs are time-dependent and can be expressed in the form:

\begin{equation}
    \frac{\partial\mathbf{u}}{\partial t} = F(t, \mathbf{x}, \mathbf{u}, \frac{\partial\mathbf{u}}{\partial x},\frac{\partial^2\mathbf{u}}{\partial x^2}, \ldots) \qquad t \in [0, T], \mathbf{x} \in \Omega
\end{equation}
where there is one time dimension \(t=[0, T]\) and multiple spatial dimensions \(\mathbf{x} = [x_1, x_2, \ldots, x_D]^T \in \Omega \), and \(\mathbf{u} : [0, T] \times \Omega \rightarrow \mathbb{R}^{d_p} \) is a quantity that is solved for with a physical dimension \(d_p\). 

Solving equations of this form is of great interest, motivating centuries of research and leading to many modern engineering and scientific advances. Although solutions are analytically intractable, they can be approximated by discretizing the domain, where \(t_n \in \{t_0, \ldots, t_N\}\), \(x_m \in \{x_0, \ldots, x_M\}\), and the solution \(\mathbf{u}(t_n, x_m)\) is discretized in this domain. Within this setup, initial conditions \(\mathbf{u}(0, x_m)\) and boundary conditions \(B[\mathbf{u}](t_n, x_b), x_b=\{x_m : x_m \in \partial\Omega\}\) are usually given based on the practical application of the PDE. 

\bmhead{Numerical Solvers} Within this setup, numerical solvers are the dominant approach to approximating PDE solutions. In general, numerical solvers aim to compute \(\mathbf{u}\) by calculating an approximation of \(\frac{\partial\mathbf{u}}{\partial t}|_{t=t_n} \approx F(t_n, \mathbf{u}(t_n))\) and evolving the current solution forward in time using this temporal derivative. This temporal derivative can be approximated by discretizing the spatial derivatives at the current time, such as using finite-difference or finite-element methods. In the most straightforward case, constructing this approximation converts the PDE into an ODE, which allows the solution to be evolved forward in time with an ODE integrator such as forward Euler or Runge-Kutta, also known as the method of lines \citep{method_of_lines}. Complex PDEs may require a more careful treatment of the temporal update, such as splitting the update into an intermediate timestep \citep{chorin_split}. As a whole, both the temporal and spatial approximation require a disciplined and informed choice in discretization schemes (explicit/implicit, 1st-order/2nd-order, triangle/quad elements, etc.); importantly, the resulting accuracy, convergence, and solution time are all influenced by these choices.

\bmhead{Neural Surrogates} As a result, deriving and solving an accurate approximation of \(\mathbf{u}\) can require significant technical expertise and computational resources. This has motivated a new class of deep learning methods, or \textit{neural surrogates}, that can learn this update from data, rather than an analytical derivation, and be quickly queried \citep{li2021fourierneuraloperatorparametric, Lu_2021}. The fast inference speed of deep learning models makes neural surrogates attractive, however, they lack performance guarantees and the rigorous mathematical theory behind numerical analysis. Regardless, data-driven neural surrogates have shown empirically good performance and the field is rapidly growing to address fundamental challenges in model accuracy, speed, robustness, and practicality. 

Many works have proposed neural surrogates based on different deep learning architectures, as well as proposed a diverse set of modifications and architecture changes to tailor networks to physics applications. These include neural surrogates based on neural operators \cite{kovachki2024neuraloperatorlearningmaps, cao2021choosetransformerfouriergalerkin}, graph neural networks (GNNs) \cite{LI2022201_graph, battaglia2016interactionnetworkslearningobjects}, transformers \cite{li2023transformerpartialdifferentialequations, alkin2024universalphysicstransformersframework, hemmasian2024multi}, convolutional neural networks (CNNs) \cite{Thuerey_2020, gupta2022multispatiotemporalscalegeneralizedpdemodeling}, or diffusion models \cite{zhou2025text2pdelatentdiffusionmodels, kohl2024benchmarkingautoregressiveconditionaldiffusion, valencia2025learning}, each with their different benefits and drawbacks. For example, CNNs are simple to train and empirically show good performance \cite{kohl2024benchmarkingautoregressiveconditionaldiffusion}, however they can only be applied to regular-grid problems. On the other hand, transformers can be used on arbitrary geometries and perform well; however, they can be costly due to the quadratic complexity of attention, and specialized algorithms have been developed to mitigate this \cite{li2023scalabletransformerpdesurrogate, wu2024transolverfasttransformersolver}. 

In addition to advancements in neural network architectures, many works have aimed to improve different aspects of neural surrogates, such as modifying the training strategy or datasets. A few examples of these works include improving the spectral performance of neural surrogates \citep{lippe2023pderefinerachievingaccuratelong}, adapting to irregularly discretized data \citep{brandstetter2023messagepassingneuralpde, li2023geometryinformedneuraloperatorlargescale}, and scaling models to large parameter sizes and across different datasets \cite{hao2024dpotautoregressivedenoisingoperator, herde2024poseidonefficientfoundationmodels, zhou2024strategiespretrainingneuraloperators, zhou2024maskedautoencoderspdelearners}. The relevant literature is too vast to detail here; however, for a list of representative neural surrogate works and their numerical baselines, we recommend \citet{McGreivy_2024} and for a survey on neural surrogates and operator learning we recommend \citet{kovachki2024neuraloperatorlearningmaps}. Within this large and continually expanding body of work, we evaluate an alternate training framework, one that can be applied generally to different architectures and can work together with different training modifications and across datasets. 

\begin{figure*}[t!]
    \centering
    \begin{subfigure}[t]{0.23\textwidth}
        \centering
        \includegraphics[width=\textwidth]{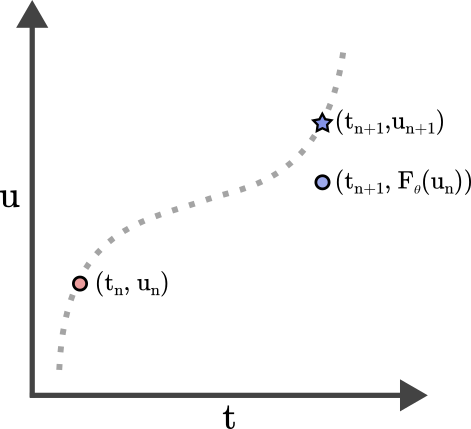}
        \caption{State prediction. Models are trained to directly predict \(\mathbf{u}_{n+1} = F_{\theta}(\mathbf{u}_n)\).}
    \end{subfigure}%
    ~ 
    \begin{subfigure}[t]{0.23\textwidth}
        \centering
        \includegraphics[width=\textwidth]{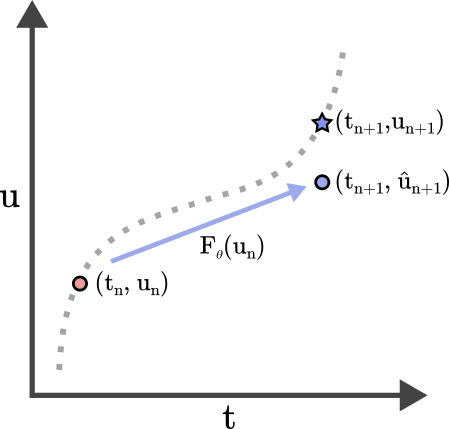}
        \caption{Derivative prediction. Models predict \(\frac{\partial\mathbf{u}}{\partial t}|_{t=t_n} = F_{\theta}(\mathbf{u}_n)\) to solve \(\hat{\mathbf{u}}_{n+1} \).}
    \end{subfigure}
    ~ 
    \begin{subfigure}[t]{0.23\textwidth}
        \centering
        \includegraphics[width=\textwidth]{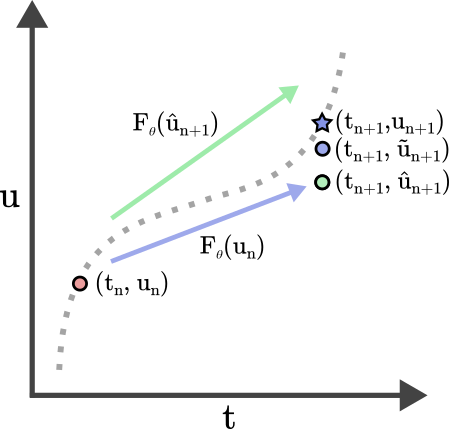}
        \caption{Derivative prediction with higher order integrators can be more accurate.}
    \end{subfigure}
    ~ 
    \begin{subfigure}[t]{0.23\textwidth}
        \centering
        \includegraphics[width=\textwidth]{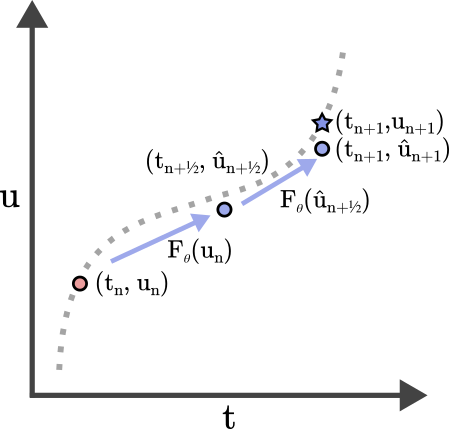}
        \caption{Derivative prediction with variable step sizes can be more flexible.}
    \end{subfigure}
    \caption{A comparison of state prediction and derivative prediction, where models are either trained to predict \(\mathbf{u}_{n+1}\) or \(\frac{\partial\mathbf{u}}{\partial t}|_{t=t_n}\). During inference, models are given an initial solution \(\mathbf{u}_n\), and predict future solutions along the dashed trajectory. By predicting the temporal derivatives rather than the future solution, derivative prediction can learn spatial updates while an ODE integrator updates the solution in time, which can improve accuracy. Furthermore, derivative prediction can use higher-order integrators or variable timesteps, which further improves its accuracy and flexibility, while being applicable across model architectures and datasets.}
    \label{fig:contribution}
\end{figure*}

\bmhead{Contributions} Neural surrogates often treat solution updates as a black box and directly predict the next state, which we call \textit{state prediction} (\(\mathbf{u}(t_{n+1}) = F_{\theta}(\mathbf{u}(t_{n}))\)). State prediction has become the dominant framework for training and querying neural surrogates, as it is the most obvious way to step forward in time, is easy to implement, and allows neural surrogates to take larger steps \(\Delta t\). However, we believe that this can be an over-simplification and, inspired by numerical methods, we ask if predicting the temporal derivative or \textit{derivative prediction}  (\(\frac{\partial\mathbf{u}}{\partial t}|_{t=t_n} = F_{\theta}(\mathbf{u}(t_{n}))\)) can improve neural surrogate accuracies. Derivative prediction changes the training objective and inference procedure, and while this framework does not contribute additional training cost, it can introduce additional inference costs. Fortunately, this is the only practical difference between the two frameworks, allowing derivative prediction to be used with any neural surrogate architecture and generally applied to the same range of PDEs, discretizations, and problem setups. 

One may ask why it is necessary to investigate alternative neural surrogate training and inference frameworks. We illustrate its differences in Figure \ref{fig:contribution} and describe its potential benefits here. Our hypothesis is that the main benefit of derivative prediction is that it allows the use of an ODE integrator to evolve the solution based on the predicted temporal derivative, which can be quickly queried at inference time. This effectively decouples the prediction problem, where the model learns the spatial dynamics that create the change in solution values, and the numerical integrator forwards the solution in time based on this learned derivative. In many PDE setups, this can greatly simplify the learning objective; rather than needing to collapse a spatial and temporal update under a single model prediction (state prediction), models can focus on learning just the change in solution values, which can be more important than the actual value of the solutions. Through experimentation, we show that this can result in more accurate model predictions. 

Secondly, derivative prediction can be more flexible during inference than state prediction. State prediction fixes the temporal resolution of the solution to the dataset resolution, whereas derivative prediction can generate solutions of arbitrary temporal resolutions by modifying \(\Delta t\) in the ODE integrator during inference, which could additionally introduce adaptive step sizing for neural surrogates. One remark is that the temporal resolution can be varied only to a point, since large \(\Delta t\) can introduce errors, especially for stiff problems. Another benefit of derivative prediction is that it can take advantage of different integration schemes, such as multi-step, predictor-corrector, or Runge-Kutta methods to improve or adjust solutions during inference. Importantly, under derivative prediction, these modifications during inference do not require re-training the model. We present experimental results demonstrating these additional capabilities.

A subtle but interesting consequence of this is that models trained with derivative prediction can use PDE data more effectively than training in the conventional state prediction framework. A common practice when training neural surrogates is to down-sample PDE data in time, which is usually finely discretized due to numerical solver constraints, and often up to factors of 10$^2$-10$^4$. This is necessary since predicting the next timestep directly implicitly fixes the temporal resolution of the learned model; training on the full-resolution data would require models to auto-regressively predict a large number of timesteps due to its fine resolution and therefore quickly accumulate error. Unfortunately, downsampling by large factors results in discarding a considerable amount of existing data, but since derivative prediction does not fix the model to a temporal resolution, this can be avoided. In particular, models can be trained on the full-resolution data to obtain highly accurate estimates of the instantaneous derivative, which can be used to take variable steps in time when numerically integrating during inference. We provide results demonstrating this as well. 

The benefits of the derivative prediction are not without drawbacks. The main limitation of derivative prediction is that it reintroduces numerical error and discretization constraints back into the neural surrogate framework. Specifically, even with perfect estimates of the instantaneous derivative, numerical integrators can still accumulate error over time with large timesteps \(\Delta t\) or stiff PDE dynamics. While this is an issue, we find that in practice, this additional numerical error is very small when compared to model error, especially when using higher-order integrators. Additionally, common timescales that conventional neural surrogates are trained on are generally small enough for ODE integrators to handle, however this is not always the case. This can be seen for steady-state problems where neural surrogates are tasked with predicting a future steady-state solution given solely the initial condition (such as Darcy Flow); derivative prediction will completely fail since the initial derivative has little correlation to the steady-state solution. While we only consider time-dependent PDEs, we seek to understand the limits of neural surrogates when predicting different timescales, both under state prediction and derivative prediction frameworks. In particular, it is known that neural surrogates can take timesteps that are numerically instable (e.g., \(CFL>1\)), but we seek to understand to what extent this is possible. Lastly, we release the code and datasets for this work here: \url{https://github.com/anthonyzhou-1/temporal_pdes}. 

\section{Related Work \label{sec:related}}
While there exists substantial prior work on using neural surrogates to model PDE solutions, our work's main contribution is proposing a method to improve solution accuracy that is agnostic to the PDE or surrogate architecture, as well as investigating an alternate training/inference framework. As such, we focus on related works that either propose surrogate-agnostic methods to improve prediction accuracies or use an alternate framework than state prediction in the model design. 

\bmhead{Neural Surrogate Modifications}
Initial work on neural surrogates designed various architectures to quickly and accurately approximate PDE solutions, including DeepONet \cite{Lu_2021}, neural operators \cite{li2021fourierneuraloperatorparametric}, and physics-informed methods \cite{RAISSI2019686}. Although powerful on their own, subsequent works have proposed modifications to these architectures to improve accuracy \cite{rahman2023unoushapedneuraloperators, li2023physicsinformedneuraloperatorlearning, wu2024transolverfasttransformersolver}, generalize to different problems \cite{hao2024dpotautoregressivedenoisingoperator, alkin2024universalphysicstransformersframework}, or adapt to complex PDE systems \cite{li2024cafaglobalweatherforecasting, pathak2022fourcastnetglobaldatadrivenhighresolution, li2022learningdissipativedynamicschaotic}. While these modifications are insightful and perform well, they are generally problem- or model-specific, which limits their applicability. To address this, prior work has also considered proposing training modifications or guidelines to improve surrogate accuracies \cite{zhou2024strategiespretrainingneuraloperators}. One such example is the use of data augmentations to improve neural surrogate training and prediction accuracy \cite{brandstetter2022lie}. In addition, previous work has explored the idea of noising data or unrolled training to improve neural surrogate stability and robustness \cite{brandstetter2023messagepassingneuralpde, hao2024dpotautoregressivedenoisingoperator, LIST2025117441}. Lastly, an interesting set of works seek to train refiner modules alongside neural surrogates to improve solution accuracy \cite{lippe2023pderefinerachievingaccuratelong}. 

\bmhead{Residual/Derivative Prediction} Prior work has also examined learning physics under different training and prediction frameworks. Besides state prediction, the most common training framework is to predict the residual between the input and the target, or \textit{residual prediction}. Residual prediction can be seen as a special case of derivative prediction, where the residual is a scaled forward Euler approximation of the temporal derivative; however, the temporal resolution is still fixed during inference. Residual prediction is used widely in weather applications \cite{lam2023graphcastlearningskillfulmediumrange, price2024gencastdiffusionbasedensembleforecasting}, and its use in PDE applications can also be found \cite{lorsung2024physicsinformedtokentransformer}. \citet{pfaff2021learningmeshbasedsimulationgraph} and \citet{sanchezgonzalez2020learningsimulatecomplexphysics} use residual prediction with GNN-based neural surrogates to approximate a Forward Euler step. \citet{stachenfeld2022learnedcoarsemodelsefficient} and \citet{wang2020physicsinformeddeeplearningturbulent} also use residual prediction with CNN-based neural surrogates, inspired by temporal updates used in numerical methods. \citet{li2022learningdissipativedynamicschaotic} investigate the effect of residual prediction as temporal resolution decreases, as a part of a larger study in learning chaotic PDE dynamics. Lastly, \citet{sanchezgonzalez2019hamiltoniangraphnetworksode} and \citet{zeng2024phympgnphysicsencodedmessagepassing} extend these ideas to use an RK2 integrator with a GNN-based architecture, after predicting the derivative using a Hamiltonian framework or with a numerical solver-inspired architecture.

Beyond PDE surrogate modeling, derivative prediction is a common framework the molecular dynamics and machine learned interatomic potential research \cite{Potential}. Rather than learning the derivative of nodal values, such as velocity, temperature, or pressure, machine learning approaches in the molecular dynamics field seek to predict the energy of a given state. This energy can then be differentiated to obtain the relevant force fields and a loss can be calculated with respect to empirical force fields from ab initio datasets \cite{force_field, force_field_2, GAMD}. 

While the idea of derivative prediction is not novel, we investigate in what situations it is beneficial, why it can help PDE surrogate models, and compare it to state-prediction and other training modifications. In addition, while other works use derivative prediction as a numerical-solver motivated technique to empirically improve surrogate models, we believe that it can be applied more generally to arbitrary models. As such, we investigate derivative prediction as an broadly applicable training framework, rather than a technique applied on an ad-hoc basis.

\bmhead{Hybrid Solvers} Another set of more restrictive, but more targeted training frameworks is to use the neural surrogate to approximate quantities within a conventional solver, creating a \textit{hybrid solver}. For example, when considering the Navier-Stokes equations, hybrid solvers can use a neural surrogate to approximate the convective flux due to its large computational burden and solve the remaining terms or steps numerically \cite{pnas_kochov, Sun_TSM2023}. For other equations, spatial derivatives can also be individually approximated using a neural surrogate \cite{pnas_bar-sinai}, and more generally, neural surrogates have found use in approximating costly terms within numerical solvers using data \cite{MARGENBERG2024116692, LIU2022104668}. Furthermore, another set of works use a numerical solver to train an inverse neural PDE solver which predicts system inputs or variables from solutions. \cite{PAKRAVAN_BiPDE, Aragon_Calvo_2020}. While these frameworks benefit from increased performance or capabilities from using a numerical solver, they are also tied to the specific problem to numerical framework (e.g, FDM, FVM, etc.) which can limit their applicability.

\bmhead{Neural Network Frameworks} While this paper considers time-dependent PDE modeling, broader interest in neural network prediction frameworks in the machine learning community has also proposed related training and inference procedures. Neural ODEs \cite{chen2019neuralordinarydifferentialequations} also predict derivatives, however these are with respect to the network's hidden state and are not used to predict physical quantities such as in derivative prediction. Additionally, this introduces significant training difficulties \cite{finlay2020trainneuralodeworld} which are not present in derivative prediction since the target is not a derivative with respect to the model parameters. If we consider the PDE to be solved under Hamiltonian or Langrangian mechanics, we can construct a neural network to predict the Hamiltonian or Lagrangian, which can be subsequently integrated to recover physical quantities continuously in time \cite{greydanus2019hamiltonianneuralnetworks, cranmer2020lagrangianneuralnetworks, lutter2019deeplagrangiannetworksusing}. Although interesting, these Hamiltonian or Lagrangian neural networks have seen little adoption to solve PDE problems largely because many PDE systems cannot be conveniently formulated with a Hamiltonian or Lagrangian, and using the approximated Hamiltonian or Lagrangian to evolve physical quantities is computationally expensive. Lastly, solving PDEs under Newtonian mechanics is more mature, with most numerical solvers and neural surrogates being developed under a Newtonian framework. 

\section{Methods \label{sec:methods}}
\subsection{Training}
During training, models parameterized by \(F_{\theta}\) are given the current solution \(\mathbf{u}(t_n)\) and timestep \(t_n\), and are trained to predict either \(\mathbf{u}(t_{n+1})\) (state prediction) or \(\frac{\partial\mathbf{u}}{\partial t}|_{t=t_n}\) (derivative prediction). This results in the following loss function \(\mathcal{L}_\theta\), for a given neural network parameterized by \(F_\theta\):
\begin{equation}
    \mathcal{L}_\theta (\mathbf{u}(t_n),t_n, \mathbf{y}) = ||F_\theta (\mathbf{u}(t_n), t_n) - \mathbf{y}||_2^2, \qquad \mathbf{y} = \begin{cases} 
      \mathbf{u}(t_{n+1}) & \text{state-prediction}\\
      \frac{\partial\mathbf{u}}{\partial t}|_{t=t_n} & \text{derivative-prediction}
   \end{cases}
\end{equation}

From a deep learning perspective, this does not change the model, architecture, or training procedure, aside from the need to compute temporal derivatives from the dataset to use as labels. This can be done numerically, such as with a forward Euler scheme (\(\frac{\mathbf{u}(t_{n+1}) - \mathbf{u}(t_{n})}{\Delta t}\)), central difference scheme (\(\frac{\mathbf{u}(t_{n+1}) - \mathbf{u}(t_{n-1})}{2\Delta t}\)), or higher-order Richardson extrapolations. Additionally, at the endpoints of the dataset a one-sided Richardson extrapolation can be used to obtain accurate estimates of the temporal derivative \cite{Kumar2006OnesidedFA}. By using increasingly higher-order finite-difference schemes, training labels can made increasingly accurate as well. 

In practice, the temporal derivatives for a given dataset can be precomputed and cached for training speed; however, even if derivatives are calculated on the fly during training, we have found the additional overhead to be small. At each training step, a random timestep \(t_n,\: n\in[0, T]\) is uniformly sampled, and the loss \(\mathcal{L}_\theta (\mathbf{u}(t_n),t_n, \mathbf{y})\) is evaluated. Additional information such as the current timestep or coefficient information is passed into the model using sinusoidal positional embeddings \cite{tancik2020fourierfeaturesletnetworks, vaswani2023attentionneed} to project the inputs to a higher dimension. Mathematically, given conditioning information \(x\in \mathbb{R
}\) (this could be \(t_n\) or coefficients \(\nu, \:c, \:\text{etc.}\)), the embedding \(\mathbf{e} \in \mathbb{R}^d\) is generated by \(\gamma: \mathbb{R} \rightarrow \mathbb{R}^d\): 

\begin{equation}
    \mathbf{e} = \gamma(x) = [\gamma(x)_1, \: \gamma(x)_2, \: \ldots, \:\gamma(x)_d]^T, \quad \text{where} \: \gamma(x)_i = \begin{cases} 
      \sin(\frac{x}{10000^{i/d}}) & i\text{ is odd}\\
      \cos(\frac{x}{10000^{i/d}}) & i\text{ is even}
   \end{cases}
\end{equation}

After the embeddings are generated, Adaptive Layer Normalization is used to condition model layers on this additional information. Specifically, we consider a neural network defined as a composition of \(L\) model layers: \(F_\theta(x) = \phi_L\circ \ldots \circ \phi_2\circ \phi_1 (\mathbf{x})\). Adaptive Layer Normalization uses a separate network at each layer \(\psi_l(\mathbf{e}): \mathbb{R}^d \rightarrow \mathbb{R}^{2d}\) to project embeddings \(\mathbf{e} \in \mathbb{R}^d\) to a scale and shift parameter \(\mathbf{\alpha} \in \mathbb{R}^{d}, \: \mathbf{\beta} \in \mathbb{R}^d\), where \(d\) is the model dimension. At a given layer \(l \in [1, 2, \ldots, L]\), the hidden activation \(\mathbf{h}_l\in \mathbb{R}^d\) is modulated by:

\begin{equation}
    \mathbf{h}_l^* = \phi_l(\mathbf{h}_{l-1}), \quad 
    [\mathbf{\alpha}, \mathbf{\beta}]^T = \psi_l(\mathbf{e}), \quad 
    \mathbf{h}_l = \mathbf{\alpha}  \mathbf{h}_l^* + \beta
\end{equation}

A final consideration is that coefficients and timesteps are scaled to be between the range \([0, 1]\) to prevent overfitting to their absolute values. Lastly, timestep information is not strictly necessary due to the temporal invariance of dynamical systems, however, including it does not change performance and the results can generalize to architectures that do not provide this information.

\subsection{Inference}
Once trained to produce accurate estimates of \(F_{\theta}(\mathbf{u}(t_n), t_n) \approx \frac{\partial\mathbf{u}}{\partial t}|_{t=t_n}\), an ODE integrator can be used to evolve the solution forward in time during inference. Specifically, we consider using the Forward Euler, Adams-Bashforth, Heun's, or 4th-order Runge-Kutta (RK4) method. The update rules are given below:

\begin{align*}
    && \mathclap{    \mathbf{u}(t_{n+1}) = \mathbf{u}(t_{n}) + \Delta t F_{\theta}(\mathbf{u}(t_{n}), t_n)} && \tag{Forward Euler} \\
    &\\
    && \mathclap{\mathbf{u}(t_{n+1}) = \mathbf{u}(t_{n}) + \frac{3\Delta t}{2} F_{\theta}(\mathbf{u}(t_{n}), t_n) - \frac{\Delta t}{2} F_{\theta}(\mathbf{u}(t_{n-1}), t_{n-1})} && \tag{Adams-Bashforth} \\
     &\\ 
    && \mathclap{\Tilde{\mathbf{u}}(t_{n+1}) = \mathbf{u}(t_{n}) + \Delta t F_{\theta}(\mathbf{u}(t_{n}), t_n)} && \\
    && \mathclap{\mathbf{u}(t_{n+1}) = \mathbf{u}(t_{n}) + \frac{\Delta t}{2}(F_{\theta}(\mathbf{u}(t_{n}), t_n) + F_{\theta}(\Tilde{\mathbf{u}}(t_{n+1}),t_{n+1})} && \tag{Heun's Method} \\
    &\\
    && \mathclap{k_1 = F_{\theta}(\mathbf{u}(t_{n}), t_n)} && \\
    && \mathclap{k_2 = F_{\theta}(\mathbf{u}(t_{n}) + \Delta t\frac{k_1}{2}, t_n + \frac{\Delta t}{2})} && \\
    && \mathclap{k_3 = F_{\theta}(\mathbf{u}(t_{n}) + \Delta t\frac{k_2}{2}, t_n + \frac{\Delta t}{2})} && \\
    && \mathclap{k_4 = F_{\theta}(\mathbf{u}(t_{n}) + \Delta tk_3, t_n + \Delta t)} && \\  
    && \mathclap{ \mathbf{u}(t_{n+1}) = \mathbf{u}(t_{n}) + \frac{\Delta t}{6}(k_1 + 2k_2 + 2k_3 + k_4)} && \tag{4th-order Runge-Kutta}
\end{align*}

When using the Forward Euler or Adams-Bashforth methods during inference, the computational cost is the same as state prediction, due to only needing to evaluate the model once per timestep. In fact, for explicit multistep methods (such as Adams-Bashforth), previous derivative estimates \( F_{\theta}(\mathbf{u}(t_{n-s}), t_{n-s})\) can be cached during inference to take more accurate forward steps without extra computational cost. However, if using Heun's or the RK4 method, the computational cost during inference is doubled or quadrupled, due to needing additional derivative estimates to evolve the solution forward in time. While more computationally expensive, this is a worthwhile tradeoff in conventional numerical schemes, as the higher accuracy of higher-order methods facilitates substantially larger timesteps \(\Delta t\), which reduces the solution time overall even when considering the additional cost of solving a single timestep. Interestingly, we show that this observation can still be true in neural surrogate settings. Lastly, the additional overhead of performing numerical integration during inference is negligible with respect to the forward pass of the model; indeed, within numerical schemes, the temporal update is usually very fast with respect to other operations.

\subsection{Models Considered}
We consider evaluating the training and inference frameworks using the Fourier Neural Operator (FNO) \cite{li2021fourierneuraloperatorparametric} and Unet \cite{gupta2022multispatiotemporalscalegeneralizedpdemodeling}, which are two popular model choices for benchmarking uniform-grid PDE prediction problems \cite{takamoto2024pdebenchextensivebenchmarkscientific, ohana2024welllargescalecollectiondiverse}. FNO models are a widely used neural operator architecture that approximates an infinite-dimensional mapping between input and output Banach spaces. This is accomplished by using the integral kernel operator \(u(x) = \int_D \kappa(x,y)v(y)\text{d}y\) to map between from an input function \(v(y)\) to an ouput function \(u(x)\). Learning the kernel \(\kappa_\theta(x,y)\) in this operator is a convenient way to parameterize an infinite-dimensional mapping; however in practice, the full integral kernel operator is expensive to both train and evaluate. While several works have seeked to mitigate this computational cost \cite{li2020neuraloperatorgraphkernel, li2020multipolegraphneuraloperator, kossaifi2023multigridtensorizedfourierneural}, the insight behind FNO models is that the integral kernel operator can be expressed as the product of the Fourier transforms of the kernel and the input function by the convolution theorem. Truncating the kernel and input Fourier series and using a Fast Fourier Transform allows the operator to be learned and evaluated quickly. In 1D, we use FNO models with \(\sim 1M\) parameters, and in 2D, we use FNO models with \(\sim7M\) parameters. We refer readers to the provided implementation for additional details on network width, depth, and number of Fourier modes. 

The theory behind FNO and neural operators as a whole guarantees an infinite-dimensional mapping \cite{kovachki2024neuraloperatorlearningmaps}; however, in practice, models relying on convolutions, such as the Unet, have shown good performance despite not satisfying discretization invariance. As such, in addition to FNO models, Unet models are a common architecture choice due to being well-studied and optimized for uniform-grid prediction tasks \cite{gupta2022multispatiotemporalscalegeneralizedpdemodeling}. Unets are widely used in the broader literature on computer vision and generative modeling \cite{ronneberger2015unetconvolutionalnetworksbiomedical, ho2020denoisingdiffusionprobabilisticmodels}, allowing many of the advances in CNN architecture design to inform PDE prediction tasks. In particular, attention blocks are placed after convolutional blocks and Wide ResNet layers \cite{zagoruyko2017wideresidualnetworks} are used. In 1D, we use Unet models with \(\sim 3M\) parameters and in 2D, we use Unet models with \(\sim 9M\) parameters. Additional details are given in the implementation for the network width, depth, and up/downsampling layers. 

\subsection{Equations Considered}
\bmhead{1D PDEs} To evaluate the proposed method, we consider testing on a wide range of PDE equations and setups. In 1D we consider the Advection, Heat, and Kuramoto-Sivashinsky (KS) equations:
\begin{flalign}
    \nonumber
    && \mathclap{\partial_t u + c \partial_x u = 0} && \text{(Adv)} \\
    \nonumber
    && \mathclap{\partial_t u - \nu\partial_{xx} u = 0} && \text{(Heat)} \\
    \nonumber
    && \mathclap{\partial_t u + u\partial_xu + \partial_{xx} u + \partial_{xxxx}u = 0} && \text{(KS)}
\end{flalign}
\noindent For all 1D equations, initial conditions are generated from a random sum of sines: 
\begin{equation}
    \label{eqn:initial_conditions}
    u(0) = \sum_{j=1}^{J}A_j sin(2\pi l_jx/L + \phi_j), 
\end{equation}
\noindent where we uniformly sample \(A_j \in [-0.5, 0.5],
 \:\omega_j \in [-0.4, 0.4], \: l_j \in \{1, 2, 3\}, \: \phi_j \in [0, 2\pi)\) while fixing \(J=5, L=16\). Samples are generated at a spatial resolution of \(n_x=100\) on a domain \(x\in[0, 16]\). The Advection and Heat equations are solved from \(t=0\) to \(t=2\), while the KS equations are solved from \(t=0\) to \(t=100\) due to the different dynamics. The first 50 seconds of the KS equation are discarded to allow the system burn-in and to converge to the chaotic state. Additionally equations coefficients are uniformly sampled for the Advection and Heat equations to introduce additional difficulty:  \(c \in [0.1, 2.5], \: \nu \in [0.1, 0.8] \). For the KS equation, coefficients are kept fixed to the more challenging dynamics. Periodic boundary conditions are used for all equations. Lastly, 4096 training samples and 256 validation samples are used for all PDEs.
 
 While conventionally a minor detail, we find that care must be taken when choosing the temporal discretization of the dataset. In state prediction, the temporal resolution must be fine enough to resolve important dynamics, yet not too fine to introduce excessive error propagation during inference. For many PDEs, such as turbulence, this can be a compromise where one objective cannot be improved without sacrificing the other \cite{lienen2024zeroturbulencegenerativemodeling}; small timescales are need to resolve the energy cascade yet this results in high autoregressive error propagation. In the context of derivative prediction, the temporal discretization is less restrictive since the model is not constrained to a resolution defined by the dataset. However, during inference, the timestep cannot be excessively large as to introduce error from the ODE integrator and shift the input distribution to the model. For the main experiments, the discretization is chosen to be 125 timesteps (\(\Delta t = 0.016s\)) for the Advection and Heat equations, and 200 timesteps (\(\Delta t = 0.25\)) for the KS equation, although these will be varied when investigating the effect of different timescales on neural surrogate performance. 

 \bmhead{2D PDEs} In 2D we consider the Burgers' and Navier-Stokes (NS) equations:
\begin{flalign}
    \nonumber && \mathclap{\partial_t u + u(\mathbf{c} \cdot \nabla u) - \nu\nabla^2u = 0} && \text{(Burgers)} \\
    \nonumber
    && \mathclap{\partial_t \omega + u \cdot \nabla \omega - \nu\nabla^2\omega = f(x), \quad  \nabla \cdot u = 0} && \text{(NS)}
\end{flalign}
 For the Burgers' equation, initial conditions are generated from a random sum of sines: 
\begin{equation}
    \label{eqn:init_2d}
    u(0) =  \sum_{j=1}^{J}A_j sin(2\pi l_{xj}x/L + 2\pi l_{yj}y/L + \phi_j)
\end{equation}
Initial condition parameters are uniformly sampled from \(A_j \in [-0.5, 0.5], \omega_j \in [-0.4, 0.4], l_{xj} \in \{1, 2, 3\}, l_{yj} \in \{1, 2, 3\}, \phi_j \in [0, 2\pi)\) while fixing \(J=5, L=2\). The vorticity form is used for the 2D Navier-Stokes equations; the initial conditions are sampled from a Gaussian random field, and the forcing function is fixed according to \citet{li2021fourierneuraloperatorparametric}. The Burgers' equation is solved on a domain \((x,y)=[-1, 1]^2\) from \(t=0\) to \(t=2\), while the Navier-Stokes equation is solved on a domain \((x,y)=[0, 1]^2\) from \(t=0\) to \(t=50\). The spatial resolution of both datasets are the same at \(64 \times 64\). The coefficients in the Burgers' equation are uniformly sampled from \(\nu \in [\num{7.5e-3}, \num{1.5e-2}]\), and \(\mathbf{c} = [c_x, c_y] \in [0.5, 1.0]^2\), while the viscosity \(\nu\) in the Navier-Stokes equations is set to \(\num{1e-3}\) for a Reynolds number of approximately \(10^3\). For the main experiments, the Burgers' equation is discretized at 100 timesteps (\(\Delta t = 0.02s\)) and the Navier-Stokes equations are discretized at 400 timesteps (\(\Delta t = 0.125s\)); additionally, the first 25\(\%\) of data for the Navier-Stokes equations (100 timesteps) are discarded to allow flow patterns to burn-in from the initial random field. 

Additionally, to evaluate performance on a more complex scenario, we consider 2D Kolmogorov flow (Kolm. Flow). Kolmogorov flow is a special case of the Navier-Stokes equation where a fixed sinusoidal forcing term is introduced along the y-axis. A drag term \(0.1\omega\) is also introduced according to \citet{pnas_kochov}. The simulation is solved using APEBench \cite{koehler2024apebenchbenchmarkautoregressiveneural} at a resolution of \(160 \times 160\) on a domain \((x,y)=[-10, 10]^2\), and the vorticity is recorded. Furthermore, the simulation is saved at a resolution of \(\Delta t = 0.1s\) for 200 timesteps, resulting in a rollout from \(t=0\) to \(t=20\) seconds. Similar to Equation \ref{eqn:init_2d}, the initial conditions are sampled from a random truncated Fourier series with 5 modes, and the viscosity \(\nu\) is set to \(\num{1e-2}\) to simulate a Reynolds number of approximately \(10^2\). Additionally, the first 25 timesteps are discarded to let the system burn-in and develop the characteristic Kolmogorov flow. Lastly, periodic boundary conditions are used and a dataset of 1024 training samples and 256 validation samples are generated for all 2D PDEs. 

\section{Results \label{sec:results}}
\begin{table}[t!]
    \centering
    \begin{tabular}{l c c c c c c}
    \toprule
         PDE: & Adv & Heat & KS & Burgers & NS & Kolm. Flow \\
         Metric: & Roll. Err. \(\downarrow\) &  Roll. Err. \(\downarrow\) & Corr. Time \(\uparrow\) & Roll. Err.  \(\downarrow\) & Roll. Err.  \(\downarrow\) & Corr. Time \(\uparrow\)\\ 
         \midrule
         FNO (State Pred.) & 0.498 & 0.589 & 139.75 & 0.437 & 0.715 & 51.4\\ 
         FNO (FwdEuler)  & 0.048& 0.141 & 79.25 & 0.196 & 0.159 & 29.0\\ 
         FNO (Adams)  & 0.032 & 0.141 & 183 & 0.174 & 0.100 & 45.5\\ 
         FNO (Heun)  & 0.032 & 0.141 & 197.75 & 0.175 & 0.100 & 81.6\\ 
         FNO (RK4)  & 0.033 & 0.141 & 198 & 0.175 & 0.100 & 82.3\\ 
         \midrule
         Unet (State Pred.) & 0.044 & 0.149 & 153.5 & 0.666 & 0.099 & 35.1\\ 
         Unet (FwdEuler) & 0.032 & 0.139 & 76.5 & 0.280 & 0.093 & 23.3\\ 
         Unet (Adams) & 0.011 & 0.139 & 167.75 & 0.264 & 0.048 & 27.9\\ 
         Unet (Heun) & 0.010 & 0.139 & 173.5 & 0.263 & 0.049 & 53.7\\ 
         Unet (RK4) & 0.010 & 0.139 & 174.5 & 0.264 & 0.049 & 88.9\\ 
         \bottomrule
    \end{tabular}
    \caption{\textbf{Prediction Accuracy.} Results on prediction accuracy across different PDEs, models, and training/inference frameworks.}
    \label{tab:pred_accuracy}
\end{table}

\subsection{Prediction Accuracy}
\label{sec:accuracy}
To compare the prediction accuracy of different training frameworks, FNO or Unet models are either trained with the state prediction objective (State Pred.), or the derivative prediction objective, where labels are calculated using a 4-th order Richardson's extrapolation. Furthermore, different integration schemes are compared during inference. Models are evaluated on the validation set by computing either the rollout error or correlation time. Rollout error is defined to be the relative L2 error of the predicted, auto-regressive trajectory \(\hat{\mathbf{u}}\) with respect to the true trajectory \(\mathbf{u}\), averaged over all timesteps, which is given by: \(\frac{1}{T}\sum_{i=1}^T\frac{||\hat{\mathbf{u}}(t_i) - \mathbf{u}(t_i)||^2}{||\mathbf{u}(t_i)||^2}\). In chaotic systems such as the KS equation or Kolmogorov Flow, rollout error can be heavily skewed and go to infinity once model predictions diverge, therefore correlation time is used to judge model performance. Correlation time is defined to be the timestep after which model predictions have a Pearson correlation of less than 0.8 with respect to the ground truth. 

Validation metrics are reported in Table \ref{tab:pred_accuracy}. We find that derivative prediction can achieve lower errors than state prediction across equations and model choices. One hypothesized reason for this is that derivative prediction can be more stable than state prediction; both models are trained such that the one-step training error is low, however, during inference, using state prediction leads to error accumulation and even unstable rollouts. Since derivative prediction simply adds a change to the previous state rather than constructing a new state, it may be able to mitigate error accumulation if the previous state is assumed to be accurate, or the timestep \(\Delta t\) is small enough such that the state change is small. 

\begin{figure}[t!]
    \centering
    \includegraphics[width=\linewidth]{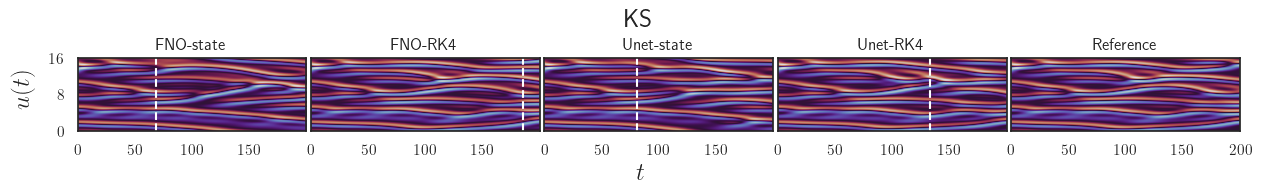}
    \caption{\textbf{1D KS Equation.} Comparison of state prediction and derivative prediction using an RK4 integrator on the 1D KS equation. Time is plotted on the x-axis and nodal values on the y-axis. Correlation time is denoted with a dashed white line.}
    \label{fig:pdes1D}
\end{figure}
\begin{figure}[t!]
    \centering
    \includegraphics[width=0.7\linewidth]{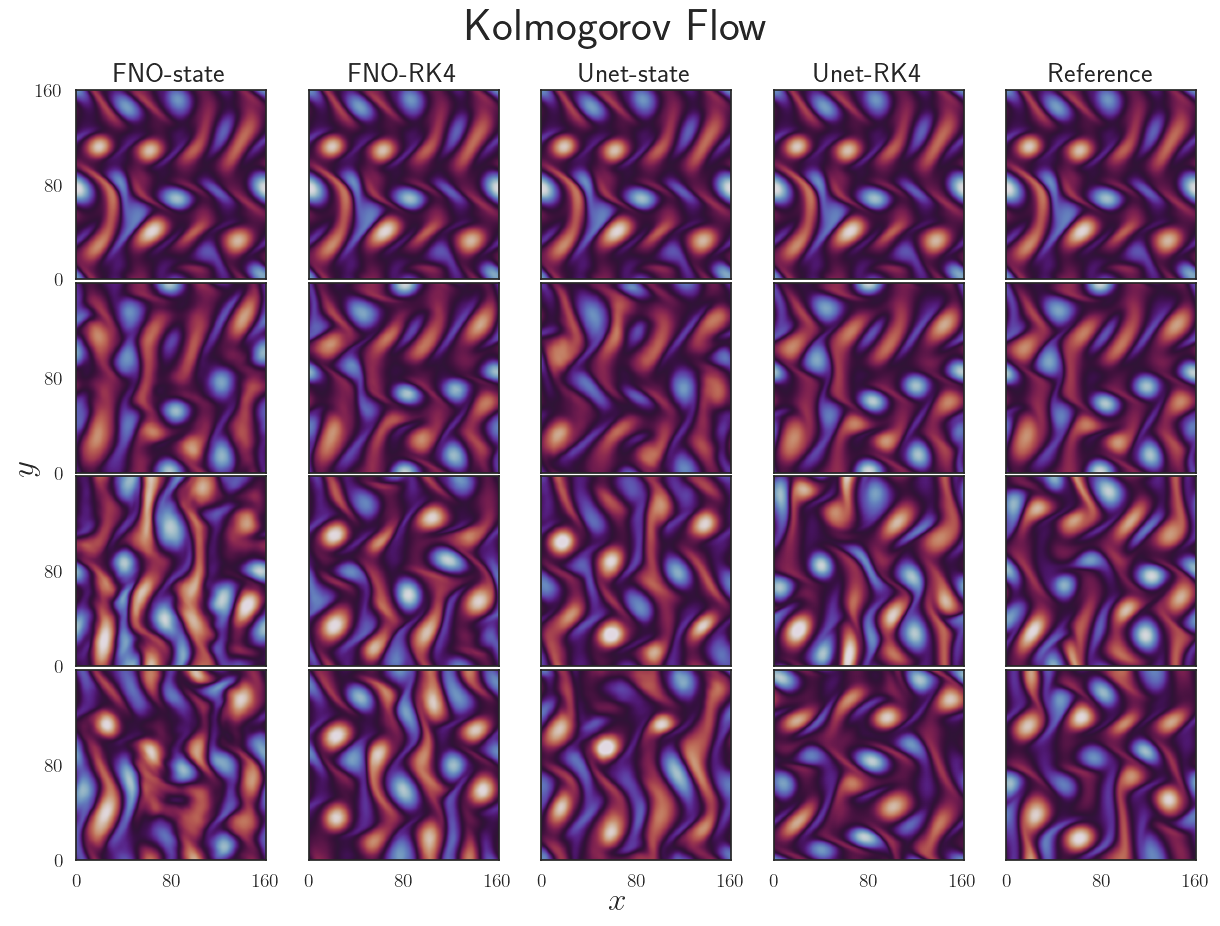}
    \caption{\textbf{2D Kolmogorov Flow.} Comparison of state prediction and derivative prediction using an RK4 integrator on 2D Kolmogorov Flow. The spatial dimensions are plotted for each frame, at multiple snapshots in time from top to bottom. The rollout is visualized at times \(0, \frac{T}{3}, \frac{2T}{3}, T\), where T is the prediction horizon.}
    \label{fig:pdes2D}
\end{figure}

For equations with simple dynamics, such as the Advection, Heat, and 2D Burgers equations, we observe that using higher-order ODE integrators does not significantly affect accuracies, although the Forward Euler method can perform worse. This doesn't mean higher-order integrators aren't useful; they can still accelerate inference times by taking larger timesteps during inference due to being more robust. However, at this resolution, the Adams-Bashforth method can be a good choice due to its higher accuracy and identical inference cost to the Forward Euler method and state prediction, due to relying on cached, past predictions. For more challenging systems, such as the KS equation or Kolmogorov flow, higher-order integrators can be more accurate, since coarser integrators can accumulate larger errors and shift the input distribution during inference. Lastly, we note that the ODE integrator can be changed during inference without re-training the model, allowing the speed or accuracy of models trained with derivative prediction to be easily changed.

Additionally, we plot sample rollouts of FNO and Unet models trained using state prediction or derivative prediction with an RK4 integrator. Results for the most challenging PDEs (KS, Kolm. Flow) are plotted in Figure \ref{fig:pdes1D} and Figure \ref{fig:pdes2D}. Qualitatively, models using derivative prediction on chaotic and complex PDE systems tend to remain accurate for longer. This can be seen directly for the KS equation, and for the Kolmogorov flow experiment, the predicted states tend to diverge at later timesteps for derivative prediction models despite all models eventually diverging.

As a whole, these results suggest that there can be benefits in learning the temporal derivative, rather than directly predicting the next state, and in practice, implementing this through different ODE integrators can further affect accuracy. Apart from being more stable, we hypothesize that learning a derivative can be beneficial due to decomposing the solution step into a spatial and temporal update; learning just the spatial update and letting an ODE integrator perform the temporal update can be easier than relying on a model to learn both. We empirically validate this in Figure \ref{fig:loss}, but leave further analysis to the discussion section.

\begin{table}[t!]
    \centering
    \begin{tabular}{l c c c c c c}
    \toprule
         PDE: & Adv & Heat & KS & Burgers & NS & Kolm. Flow\\
         Metric: & Roll. Err. \(\downarrow\) &  Roll. Err. \(\downarrow\) & Corr. Time \(\uparrow\) & Roll. Err.  \(\downarrow\) & Roll. Err. \(\downarrow\) & Corr. Time \(\uparrow\)\\ 
         \midrule
         FNO (State Pred.) & 0.498 & 0.589 & 139.75 & 0.437 & 0.715 & 51.4\\ 
         FNO (4x params)  & 0.826 & 0.991 & 147.75 & 0.276 & 0.599 & 67.4\\ 
         FNO (Pushforward)  & 0.357 & 0.486 & 141 & 0.397 & 0.090 & 53.3\\ 
         FNO (Refiner)  & 0.036 & 0.167 & 159.75 & 1.141 & 0.596 & 37.5 \\ 
         FNO (RK4)  & 0.033 & 0.141 & 197.25 & 0.175 & 0.100 & 82.3\\ 
          FNO (RK4+Pushforward)  & 0.023 & 0.140 & 194.25 & 0.322 & 0.058 & 83.8 \\ 
         \midrule
         Unet (State Pred.) & 0.044 & 0.149 & 153.5 & 0.666 & 0.099 & 35.1\\ 
         Unet (4x params) & 0.036 & 0.144 & 145.25 & 0.222 & 0.053 & 52.1\\ 
         Unet (Pushforward) & 0.048 & 0.149 & 145 & 0.746 & 0.078 & 36.0\\ 
         Unet (Refiner) & 0.082 & 0.176 & 145.5 & 0.217 & 0.180 & 51.1\\ 
          Unet (RK4) & 0.010 & 0.139 & 174.5 & 0.264 & 0.049 & 88.9\\ 
        Unet (RK4+Pushforward) & 0.010 & 0.139 & 177 & 0.264 & 0.034 & 90.6 \\ 
         \bottomrule
    \end{tabular}
    \caption{\textbf{Training Modifiers.} Comparing different training modifications on various PDEs, models and frameworks.}
    \label{tab:modifiers}
\end{table}

\subsection{Comparison to Training Modifiers}

We compare our approach with various methods to improve the prediction accuracy of models, including using 4 times as many model parameters, implementing the pushforward trick/unrolled training \cite{brandstetter2023messagepassingneuralpde, LIST2025117441}, or using the PDE-Refiner framework \cite{lippe2023pderefinerachievingaccuratelong}. The pushforward trick, or unrolled training, aims to improve rollout stability and accuracy by introducing model noise during training. This is done by making a model prediction \(\hat{\mathbf{u}}\), then using this model prediction as an input to make another prediction \(F_{\theta}(\hat{\mathbf{u}})\), which is compared to the ground truth to evaluate a loss. This unrolling can also be done for multiple timesteps, however, the compute and memory scales linearly with each additional unrolled timestep. Additionally, the gradient can be calculated with respect to only the final timestep or all timesteps (non-differentiable/differentiable unrolling). For our implementation, we follow \citet{brandstetter2023messagepassingneuralpde} and unroll the training for 1 additional timestep while letting this unrolling be non-differentiable. Furthermore, we allow the model to train without the pushforward trick for a certain number of initial epochs to prevent excessive model noise from degrading its training. 

While unrolled training is a simple and often effective method to improve neural surrogate accuracy, further gains in accuracy have been demonstrated by using a probabilistic refinement of predicted solutions to better recover the true solution spectrum, which we call refiner methods \cite{lippe2023pderefinerachievingaccuratelong, oommen2024integratingneuraloperatorsdiffusion}. These methods work by adding Gaussian noise to the prediction target and training a refinement model to denoise this input to match the target. While the specific implementation can vary, these methods generally sample random noise during inference and, conditioned on a past timestep, take several refinement steps to produce a future solution. For our benchmark, we follow \citet{lippe2023pderefinerachievingaccuratelong} and use three refinement steps. 

We train models using various modifications with the state prediction framework and additionally report baseline models using only state prediction and derivative prediction with an RK4 integrator. Validation metrics are reported in Table \ref{tab:modifiers}. We reproduce previous results showing the effectiveness of scaling model size, unrolled training, and refiner methods, which tend to improve the performance of neural surrogates across many PDEs and models. However, we observe the improvement of these modifications can be more nuanced. For example, scaling the model size can depend on the architecture's scalability and capacity, as well as data constraints to prevent overfitting. Indeed, we observe that FNO models tend to overfit in simple test cases (Advection, Heat), while Unet models do not, underlining previous results on the scalability of Unet vs. FNO models in PDE contexts \cite{zhou2024strategiespretrainingneuraloperators}. Furthermore, while refiner methods are very powerful, in some cases achieving the best performance, introducing a denoising objective creates additional complexity during training, and in certain cases performs worse than a baseline. Refiner methods can also be dependent on model architecture, with some models performing better than others on denoising; in particular, FNO models can struggle with spurious high frequencies introduced by Gaussian noise that are exacerbated in 2D PDEs. As a whole, derivative prediction can be a useful tool to improve model performance and can have improvements comparable or better than other training modifications. In addition, derivative prediction does not add additional training cost, while other training modifications generally require more compute and memory during training.

It is also interesting to note that these modifications are not exclusive of each other; in fact, it is straightforward to implement derivative prediction in conjunction with scaling the model size, using unrolled training, or applying refiner methods to further improve performance. We demonstrate this by training a model using derivative prediction (RK4) with the pushforward trick; the model makes a prediction of the current derivative \(F_\theta(\mathbf{u}(t)) = \frac{d\mathbf{u}}{dt}|_{t=t}\), it is integrated to obtain the next predicted solution \(\hat{\mathbf{u}}(t+1)\), and this is used to predict the following derivative \(F_\theta(\hat{\mathbf{u}}(t+1)) = \frac{d\mathbf{u}}{dt}|_{t=t+1}\), which is finally used in the loss function. In simple cases, we find that the pushforward trick is unnecessary, however for more complex PDEs, combining this modification with derivative prediction can have modest improvements. 

As a final remark, while overall performance gains of these methods support prior work and observations, applying these training modifications successfully still requires an understanding of the problem setup, model capacity, dataset size, or relevant theory, as the results show that training intricacies and exceptions to overall trends still exist. Much like any numerical or machine-learning method, using derivative prediction similarly has these considerations, however, we leave this discussion to later sections.

\begin{table}[t!]
    \centering
    \begin{tabular}{l c c | l c c }
    \toprule
         PDE: & Adv & NS & PDE: & Adv & NS \\
         Metric: & Roll. Err. \(\downarrow\) &  Roll. Err. \(\downarrow\) & Metric: & Roll. Err. \(\downarrow\) &  Roll. Err. \(\downarrow\)\\ 
         \midrule
         FNO (FwdEuler) & 0.048  & 0.159 & Unet (FwdEuler) & 0.032  &  0.093\\ 
         FNO (FwdEuler + 2x data)  & 0.045 & 0.139 & Unet (FwdEuler + 2x data) & 0.033 & 0.091\\ 
         FNO (FwdEuler + 2x steps) & 0.037 & 0.120 & Unet (FwdEuler + 2x steps) & 0.019  & 0.059\\ 
         FNO (Heun) & 0.033 & 0.100 & Unet (Heun) & 0.010 & 0.049\\ 
         \bottomrule
    \end{tabular}
    \caption{\textbf{Inference Modifiers.} Comparing different inference modifications across various models and PDEs.}
    \label{tab:inference}
\end{table}

\subsection{Inference Modifications}

A potential benefit of derivative prediction is its flexibility: predicting temporal derivatives allows the temporal resolution of the model to not be fixed. This can be taken advantage of to train on more data, not in the sense of generating more data samples but using more of the existing data at a finer temporal discretization. In fact, PDE data are usually heavily downsampled in time to prevent models that use state prediction from being fixed to a fine temporal resolution, yet this discards a large proportion of the generated data. We test if using more data is beneficial by training a model using derivative prediction on data discretized at half the temporal resolution (2x data). For our test cases, this results in using 250 timesteps instead of 125 timesteps for Adv/Heat and 600 timesteps instead of 300 timesteps for NS, while keeping the start and end time of the trajectories the same. During inference, models still make predictions at the native resolution, after being trained on twice the data at the finer resolution.

Alternatively, since the temporal resolution is not fixed, during inference the ODE integrator can take smaller steps in time to potentially improve neural surrogate accuracies. We test whether taking more steps is beneficial by training a model at the native resolution, but taking twice as many steps (2x steps) during inference. This amounts to setting \(\Delta t\) to half its original size in the ODE integrator. Additionally, predictions at half-steps are discarded, and rollout errors are calculated at the native resolution.

We report the results of training on more finely discretized data (2x data) and taking smaller timesteps during inference (2x steps) in Table \ref{tab:inference}. Results are evaluated for the Advection and Navier-Stokes equations to validate these ideas on a simple 1D case and a more complex 2D case. In addition, we also report baselines using the Forward Euler or Heun's method without modifications. Interestingly, training on more finely discretized data can result in lower rollout errors without additional inference cost, likely from learning more accurate derivative estimates. For simple systems (Adv), this additional data does not seem to be necessary, but for more complex systems (NS), more data and better derivative estimates become more beneficial. This opens opportunities for models trained using derivative prediction to discard less data than conventional state prediction frameworks and achieve better performance. However, we hypothesize this can have diminishing returns, as further increasing the resolution of the dataset results in adding data samples that are already very similar to the existing dataset. 

Furthermore, taking smaller steps during inference can also decrease rollout error, but at the cost of more compute. Interestingly, taking more steps does not result in higher error propagation; despite rolling out trajectories to twice the original length (even up to 600 steps in NS), the error remains stable and even decreases when the trajectory length increases. This is opposite to conventional observations when training models using state prediction, where rollout error increases as more steps are taken during inference \cite{brandstetter2023messagepassingneuralpde, lienen2024zeroturbulencegenerativemodeling, lippe2023pderefinerachievingaccuratelong}. Again, we hypothesize that this stability is due to using a numerical integrator to evolve forward in time, rather than relying on the model to do so. 

However, we note that lower errors can also be achieved by using higher-order integrators rather than smaller steps. Indeed, Heun's method adds a corrector step to the Forward Euler method to achieve a higher-order approximation, but uses twice the compute. Therefore, while the computational cost of using Heun's method and the Forward Euler method with half the step size is equal, using Heun's method during inference is still more accurate. A consequence of this is that if speed is preferred during inference, using higher-order schemes with a larger step size could still outperform lower-order schemes with smaller step sizes, despite the additional compute of higher-order schemes. Alternatively, in cases where additional accuracy is important, higher-order integrators can also be used with smaller step sizes to improve accuracy. Indeed, one of the main benefits of derivative prediction is being able to adaptively change the step size during inference, which can lead to taking larger steps where the solution changes slowly and smaller steps where the solution changes quickly. 

\begin{figure*}[t!]
    \centering
    \begin{subfigure}[t]{0.5\textwidth}
        \centering
        \includegraphics[width=\textwidth]{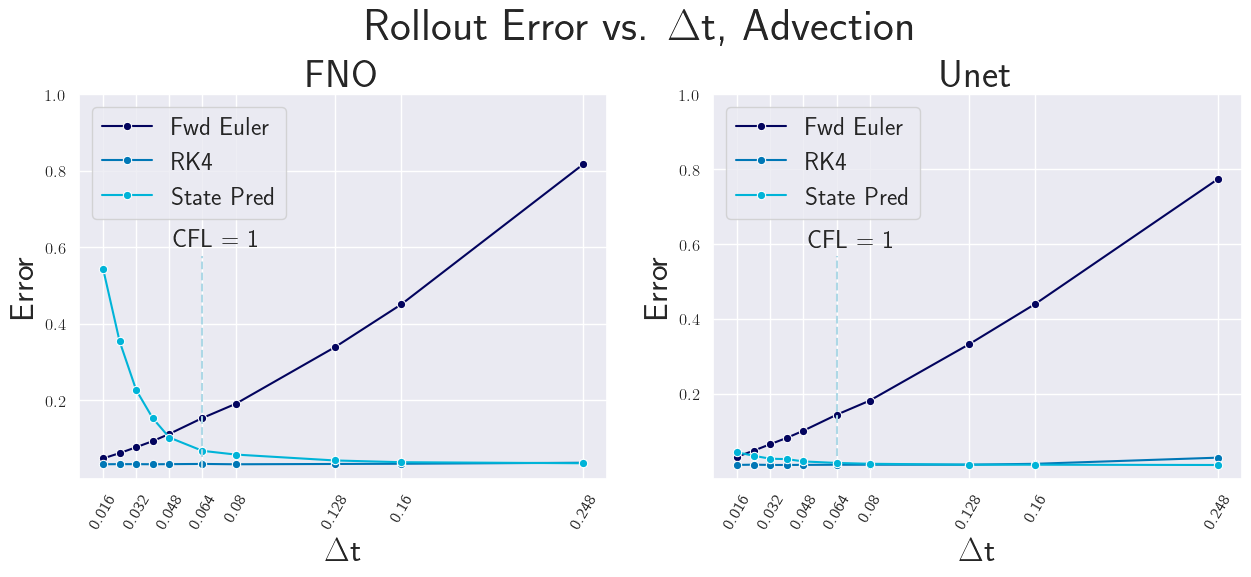}
        \caption{1D Advection equation.}
    \end{subfigure}%
    ~ 
    \begin{subfigure}[t]{0.5\textwidth}
        \centering
        \includegraphics[width=\textwidth]{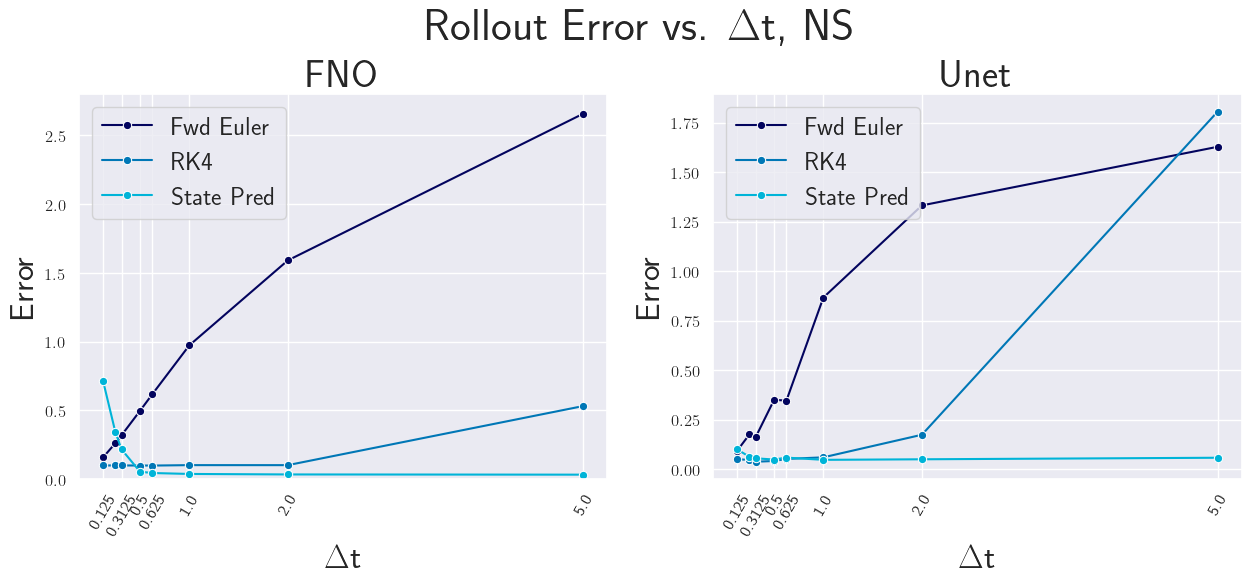}
        \caption{2D Navier-Stokes equation.}
    \end{subfigure}
    \caption{\textbf{Prediction Timescales.} Rollout error of FNO or Unet models trained with state prediction or derivative prediction with a Forward Euler or RK4 integrator; errors are plotted after predicting solutions at different timescales \(\Delta t\). The resolution at which \(CFL=1\) is denoted for 1D Advection. }
    \label{fig:timescales}
\end{figure*}

\subsection{Prediction Timescales} 
The main limitation of derivative prediction is that it reintroduces discretization constraints and numerical error to neural surrogates. To evaluate the extent of this limitation, we consider training models using state prediction at different temporal resolutions, as well as evaluating a model trained using derivative prediction at different step sizes \(\Delta t\). This is done for both FNO and Unet models, and with the 1D Advection and 2D Navier-Stokes equations. For 1D Advection, the different timescales considered are: \(\Delta t = 0.016, \: 0.024, \:0.032, \:0.04, \:0.048, \:0.064, \:0.08, \:0.128,\:0.160, \:0.248\) seconds, corresponding to predicting 125, 80, 60, 50, 40, 30, 25, 15, 12 or 8 steps of the solution trajectory. For the 2D Navier-Stokes equations, the different timescales considered are: \(\Delta t = 0.125, \: 0.25, \:0.3125, \:0.5, \:0.625, \:1.0, \:2.0, \:5.0\) seconds, corresponding to predicting 400, 200, 150, 100, 75, 50, 25, or 10 steps of the solution trajectory. One remark is that models using state prediction need to be re-trained for each resolution, while a single model trained using derivative prediction can be queried at these resolutions during inference.

The results of this evaluation are plotted in Figure \ref{fig:timescales}. We observe that as \(\Delta t\) increases during inference, models trained with derivative prediction tend to have higher errors due to accumulating error from numerical integration. Using higher-order schemes, such as RK4, alleviates this problem, but for more complex systems (NS) the step size can eventually become too large. We observe that the opposite trend is true for models trained with state prediction. In fact, it is well known that the rollout error of models decreases as the trajectory length decreases (i.e., \(\Delta t\) increases), due to lower error propagation and the ability for neural surrogates to circumvent conventional step sizing constraints. 

Indeed, for the 1D Advection equation, models trained with state prediction remain stable in regimes where the CFL number (\(\frac{c\Delta t}{\Delta x}\)) is well above 1. Interestingly, this is also true for models trained with derivative prediction if a suitable integrator is used during inference. For the 2D Navier-Stokes equations, a similar stability criterion does not exist, however a very coarse criterion can be estimated by considering the stability of just the viscous term \(\nu \nabla^2\mathbf{u}\). This viscous term is stable when \(\Delta t \leq \frac{(\Delta x)^2}{2\nu} \approx 0.122s\), which gives an upper bound to the step sizing; in reality, non-linear terms in the Navier-Stokes equations will likely necessitate a much smaller step size. Therefore, all timescales considered for the Navier-Stokes equations are numerically unstable, yet models trained with both state prediction and derivative prediction can still be accurate, albeit with different exceptions. State prediction can be unstable at small \(\Delta t\) due to auto-regressive error accumulation and derivative prediction can be unstable at large \(\Delta t\) due to error from numerical integration.

Given these limitations of derivative prediction it is worth addressing to what extent this is an issue. Despite including a numerical integrator, derivative prediction can still be stable in highly numerically unstable regimes due to having data-driven derivative estimates. However, at very large \(\Delta t\) this is no longer true; however, predictions with such coarse temporal resolution may not even be useful. In fact, when taking only 10 steps for the Navier-Stokes equations (\(\Delta t = 5s\)) the predicted frames are so distinct that meaningful dynamics cannot be reconstructed from these distant frames. Compared to state prediction, using derivative prediction can require taking more steps during inference to maintain stability, however, this can still be fewer steps than a numerical solver. For example, the pseudo-spectral method used to solve the Navier-Stokes equations uses \(\Delta t = \num{1e-4}s\), which corresponds to taking 500,000 steps; with more optimization this can be solved at a coarser resolution, however, derivative prediction will likely still use fewer steps. 

\begin{figure*}[t!]
    \centering
    \begin{subfigure}[t]{0.5\textwidth}
        \centering
        \includegraphics[height=1.3in]{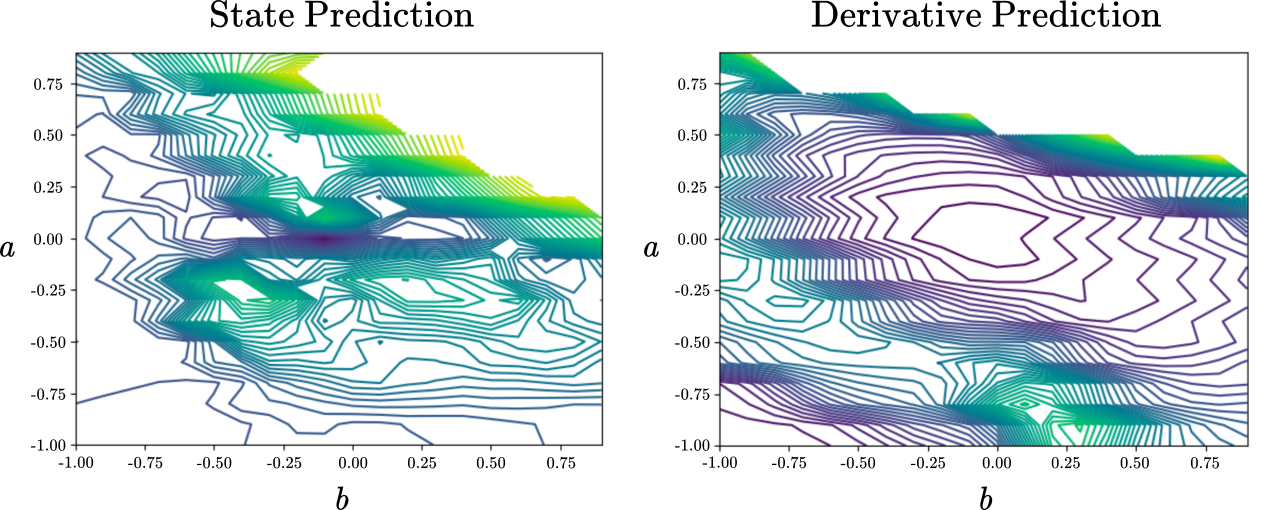}
        \caption{Loss contours of state prediction (left) and derivative prediction (right)}
    \end{subfigure}%
    ~ 
    \begin{subfigure}[t]{0.5\textwidth}
        \centering
        \includegraphics[height=1.3in]{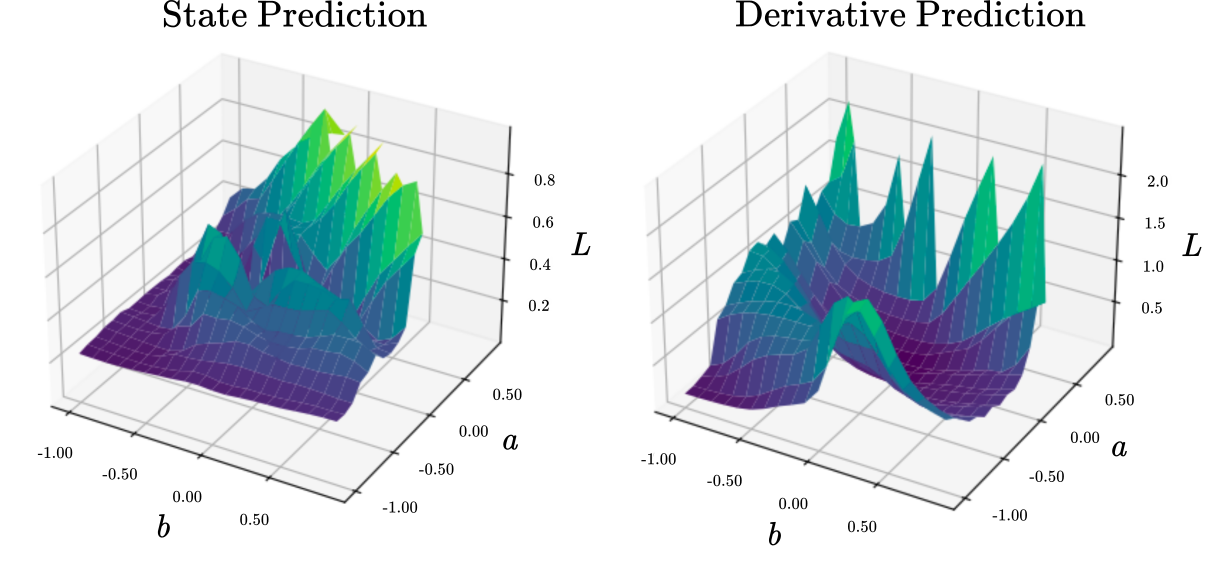}
        \caption{Loss surfaces of state prediction (left) and derivative prediction (right)}
    \end{subfigure}
    \caption{\textbf{Loss Landscapes.} Visualization of the loss landscape of state prediction and derivative prediction after training a Unet on 1D Advection. Trained model parameters \(\theta^*\) are varied by a linear combination of random direction vectors \(\delta,\gamma\) at sampled \(a, b\) values. Validation rollout loss for each perturbed set of weights \(\theta^* + a\delta + b\gamma\) are calculated and averaged across the validation set, which is plotted as contours or surfaces. The scale and threshold of \(L\) is kept consistent across contour and surface plots for a given model/framework.}
    \label{fig:loss}
\end{figure*}

\section{Discussion \label{sec:discussion}}

\subsection{Loss Landscapes}

To understand why derivative prediction may outperform state prediction, we train a Unet on the 1D Advection benchmark using both state and derivative prediction, and visualize the loss landscape using filter normalization \cite{li2018visualizinglosslandscapeneural} in Figure \ref{fig:loss}. The loss landscape is constructed by perturbing trained weights along a linear combination of normalized, random directions and evaluating the perturbed rollout loss on validation samples. Specifically, given \(\theta^* \in \mathbb{R}^n\), which are the trained model parameters (n is the parameter count), we define two random vectors \(\delta \in \mathbb{R}^n\) and \(\gamma \in \mathbb{R}^n\). Given a loss function \(\mathcal{L}_\theta (x,y)\), multiple rollout loss values are calculated after varying the network’s parameters \(\mathcal{L}_{\theta^* + a\delta + b\gamma} (x, y)\) by a linear combination of \(\delta\) and \(\gamma\). Each time a loss is calculated with a specific set of parameters \(\theta^* + a\delta + b\gamma\), it is averaged over the validation set \((x,y) \in D_{val}\). We sample \(a\) and \(b\) along a grid, which form the two axes, with \(\mathcal{L}_{\theta^* + a\delta + b\gamma} (x, y)\) being the third, vertical axis or contour set.

\begin{figure}[t!]
    \centering
    \includegraphics[width=0.8\linewidth]{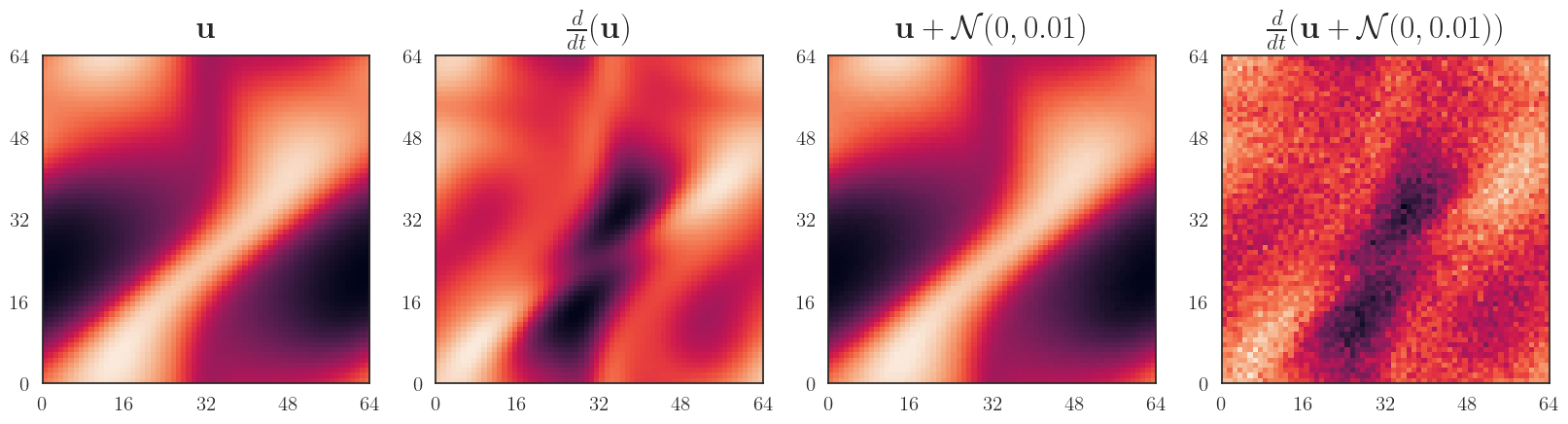}
    \caption{\textbf{Noised Trajectories.} Snapshots of the 2D Navier-Stokes equation, along with their temporal derivatives. Adding noise to a trajectory creates a similar sample, however this effect is noticeable when taking its temporal derivative.}
    \label{fig:noise}
\end{figure}

We find that the derivative prediction objective can be empirically easier to learn than the state prediction objective. One potential reason for this is that when solving PDEs at a suitable resolution, future timesteps are often similar to current ones but with slight changes from evolving forward in time. This causes the prediction label to contain redundant information about the current state, where the learning signal comes from the change in state, which can be small. This can contribute to many local minima, where models can learn variations of the identity map that predict potential states that resemble the ground truth, since the forward evolution only slightly changes the solution \cite{li2022learningdissipativedynamicschaotic}. However, despite nodal values being similar among local minima, the temporal derivatives at these potential states can be very different. Therefore, derivative prediction can isolate the important dynamics to better distinguish between potential model predictions. 

One way to visualize this phenomena is to compare ground-truth and noised PDE data and their derivatives. This is plotted in Figure \ref{fig:noise}. A ground-truth trajectory \(\mathbf{u}\) at a given time \(t\) is plotted along with the derivative evaluated at that time \(\frac{\partial}{\partial t}(\mathbf{u})\). A small amount of Gaussian noise (\(\mathcal{N}(0, 0.01)\)), which is centered at \(\mu = 0\) and with variance \(\sigma^2 = 0.01\), is added to \(\mathbf{u}\) at all timesteps, and the derivative is evaluated with respect to this noised trajectory; these quantities are also plotted at the given time \(t\). We make a similar observation: many local minima can be similar to the true state, and indeed adding a small amount of Gaussian noise nearly resembles the true label; however, when taking a temporal derivative, differences between local minima and the true label are highlighted to help models learn the true solution.

\subsection{Error Analysis}
After investigating the performance of different training frameworks across models, PDEs, training modifications, and timescales, we are also interested in examining the sources of error to better understand the mechanisms behind derivative prediction. 

\begin{table}[t!]
    \centering
    \captionsetup{width=16cm}
    \begin{tabular}{l c c c c c c}
    \toprule
         Model & Adv & Heat & KS & Burgers & NS & Kolm. Flow \\
         \midrule
         FNO-State (Next-Step Error) & 0.0048 & 0.0053 & 0.0012 & 0.0382 & 0.0016 & 0.0214\\ 
         FNO-Derivative (Next-Step Error) & 0.0004 & 0.0020 & 0.0003 & 0.0045 & 0.0005 & 0.0038\\ 
         FNO-Derivative (Derivative Error) & 0.0204 & 0.2665 & 0.0047 & 0.1104 & 0.0234 & 0.0522\\
         \midrule
         Unet-State (Next-Step Error) & 0.0041 & 0.0044 & 0.0013 & 0.0291 & 0.0018 & 0.0254 \\ 
         Unet-Derivative (Next-Step Error) & 0.0003 & 0.0018 & 0.0008 & 0.0068 & 0.0008 & 0.0094 \\ 
         Unet-Derivative (Derivative Error) & 0.0101 & 0.3065 & 0.0142 & 0.1700 & 0.0402 & 0.1256 \\
         \bottomrule
    \end{tabular}
    \caption{\textbf{Next-Step Error.} Results on validation next-step error across different PDEs, models, and training/inference frameworks. Derivative error is calculated with an RK4 integrator. Next-step error for state-prediction models is given by: \(\mathcal{L}(\mathbf{u}(t_{n+1}), F_\theta(\mathbf{u}(t_n)))\), while for derivative-prediction models both the derivative error: \(\mathcal{L}(\frac{d\mathbf{u}}{dt}|_{t=t_n}, F_\theta(\mathbf{u}(t_n)))\) and next-step error: \(\mathcal{L}(\mathbf{u}(t_{n+1}), \int F_\theta(\mathbf{u}(t_n)))\) are given.}
    \label{tab:next-step}
\end{table}

\bmhead{Next-Step Error}
While rollout error and correlation time are presented in the main results, examining next-step error can give information into how well the model is able to learn its training objective. While the transition from next-step error to rollout error is straightforward in state-prediction (rollout error is the sum of next-step error along a trajectory), it is worthwhile to consider the next-step error for derivative-prediction. Specifically, we are interested in how well models can match empirical derivatives and if integrating the learned derivative gives a similar next-step error to state-prediction frameworks. We present these results in Table \ref{tab:next-step} for all models and PDEs. Next-step error is reported for state-prediction models and derivative-prediction models, averaged over all timesteps in the validation set. For state prediction models, next-step error is given by: \(\mathcal{L}(\mathbf{u}(t_{n+1}), F_\theta(\mathbf{u}(t_n)))\), while for derivative-prediction models we report both derivative error: \(\mathcal{L}(\frac{d\mathbf{u}}{dt}|_{t=t_n}, F_\theta(\mathbf{u}(t_n)))\) and next-step error: \(\mathcal{L}(\mathbf{u}(t_{n+1}), \int F_\theta(\mathbf{u}(t_n)))\) after integrating the predicted derivative. Furthermore, to compare the losses between the datasets and the models, all errors are calculated using a relative L2 loss, as in Section \ref{sec:accuracy}.  

After decoupling the derivative prediction and ODE integrator, we find that the derivative error is consistently higher than next-step error, yet it is still low enough to achieve better performance than state prediction models. We hypothesize this is because temporal derivatives can be more challenging to predict than solution fields; for example, in Burgers equation, nonlinear terms can cause sharp discontinuities in \(\frac{d\mathbf{u}}{dt}\). However, using a sufficiently fine temporal resolution and a higher-order integrator can account for this error by scaling the derivative by \(\Delta t\) and using multiple derivative predictions. Both the model and time-stepping scheme can contribute to lower next-step error, and subsequently lower rollout error, for derivative prediction models. 

\bmhead{Numerical Error}
While models can learn accurate derivative estimates, this is not enough to maintain accuracy over a rollout. In particular, even with access to perfect derivatives, numerical integrators can still accumulate discretization error. Therefore, we seek to understand to what extent this affects derivative prediction rollouts. In particular, we have access to ground-truth derivative estimates from the ground-truth trajectory; using these estimates rather than neural surrogate estimates can serve as a baseline to compare how much error is being contributed by the ODE integrator. This baseline is called the numerical \textit{oracle}, since it relies on future states to accurately estimate the instantaneous derivative (i.e., when given an initial state, we look into the future to compute \(\frac{d\mathbf{u}}{dt}\)). 

\begin{figure}[t!]
    \centering
    \includegraphics[width=0.7\linewidth]{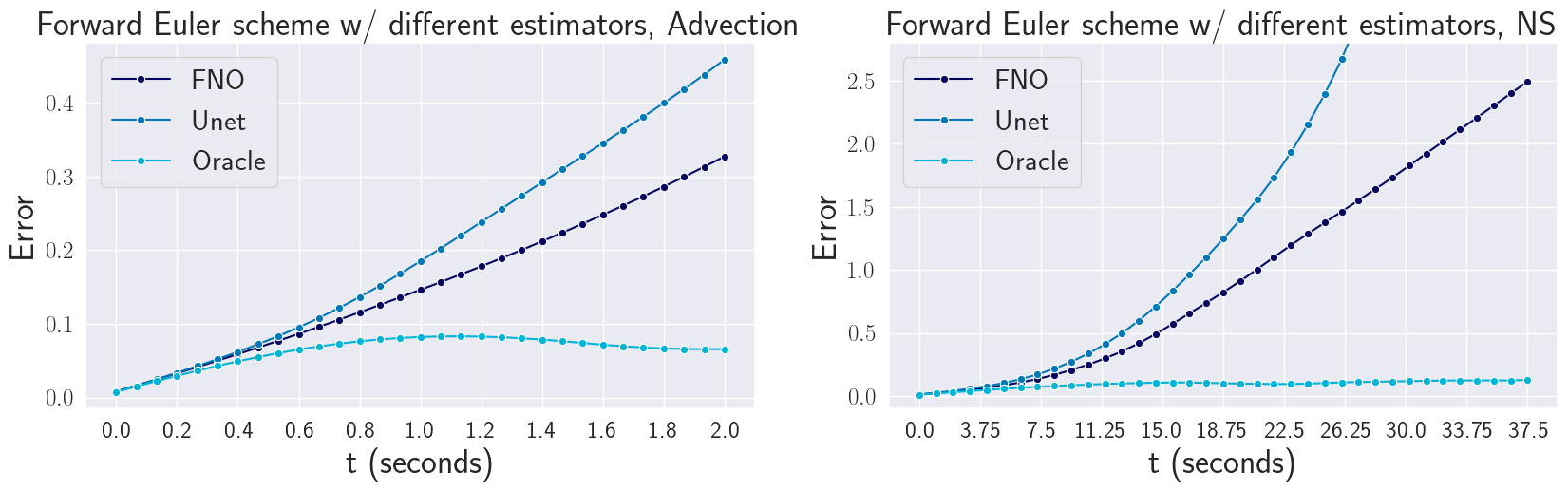}
    \caption{\textbf{Numerical Error.} Error at each timestep using FNO, Unet, or an oracle to estimate \(\frac{\partial \mathbf{u}}{\partial t}\) and integrating with the Forward Euler method, for the 1D Advection and 2D NS equations. Each marker denotes an integration step.}
    \label{fig:numerical}
\end{figure}

We plot the error over time during inference on the validation set of the 1D Advection and 2D NS equations in Figure \ref{fig:numerical}. The solution is integrated at a sufficiently coarse step size to introduce instability (\(\Delta t = 0.064s\) for advection and \(\Delta t = 1s\) for NS). The ODE integrator, which in this case is the Forward Euler method, is given derivatives from trained FNO or Unet models, as well as the oracle. We find that despite having near-perfect derivative estimates from the oracle, the integrator still accumulates error over time. However, these errors are still much smaller than querying a neural surrogate for derivative estimates. Therefore, we conclude that the integrator can introduce numerical error, however it is very small with respect to model error, even when considering numerically unstable step sizes. In practice, this means that the model, rather than the framework, is the primary opportunity for improvement. 

From the previous analysis, we know that the next-step validation loss is low and can assume that if given a perfect input, models can make an accurate derivative estimate. Indeed, this seems to be the case where the model errors closely follow the oracle errors up to a certain point (around the first \(20\%\) of the rollout). However, after this point, the numerical error shifts model inputs such that they no longer match the training distribution, and the model begins to make inaccurate derivative estimates, which contributes model error. Integrating these inaccurate estimates produces model inputs even further from the training distribution, which produces even more inaccurate derivative estimates. This error compounds until the model no longer makes useful predictions. 

One avenue to delay this error propagation is to use higher-order schemes to reduce the numerical error and delay the initial shift in model inputs. In fact, this is one reason why higher-order schemes remain more stable at larger time steps as shown in Figure \ref{fig:timescales}. However, reducing model error and improving noise robustness can also delay this error propagation; this opens up promising avenues to apply unrolled training or refiner methods with derivative prediction, which are designed for this purpose.

\subsection{Timing Experiments}
A final consideration for derivative prediction is that it can take longer during inference, since higher-order integrators can require multiple predictions per step. Therefore, we compare the computational cost of models using derivative prediction to those using state prediction, as well as to baseline numerical solvers. The results are given in Table \ref{tab:timing} for the Navier-Stokes and Kolmogorov Flow experiment. Models are run during inference to predict a full rollout for each validation sample, and times are averaged across the validation set. Furthermore, model sizes are kept consistent across all timing experiments. We use the pseudospectral solver for the Navier-Stokes experiment proposed by \citet{li2021fourierneuraloperatorparametric} and a JAX-implemented, pseudospectral ETDRK4 solver for the Kolmogorov Flow experiments, proposed by \citet{koehler2024apebenchbenchmarkautoregressiveneural}. Batch sizes are set to maximize GPU usage across all experiments. Following \citet{McGreivy_2024}, we also coarsen the solver resolution to match the neural surrogate accuracy when needed to provide a fair comparison of computational cost. The rollout accuracy of the coarse solution is calculated by using bilinear interpolation to map the coarse solution onto the finer grid, and the relative L2 error is calculated for the interpolated trajectory and averaged across the validation set. All computations are performed on a single NVIDIA RTX 2080 Ti GPU for the Navier-Stokes experiment and a single NVIDIA RTX 6000 Ada GPU for the Kolmogorov Flow experiment. 

We find that with lower-order schemes, the computational cost of derivative prediction is not meaningfully higher than state prediction. This is because in the Forward Euler scheme, the additional overhead is extremely small, as we only need to add and scale the predicted derivative to the current value. Likewise, for the Adams-Bashforth scheme, despite using multiple model predictions and having a lower truncation error, the previous predictions are cached and therefore do not need to be re-evaluated, resulting in minimal overhead. However, for Heun's method and the RK4 integrator, model predictions are needed to correct the predicted step or predict derivatives at interpolated timesteps. These additional model evaluations roughly double or quadruple the inference cost, which is consistent with the amount of extra derivatives that need to be calculated in the numerical schemes. Another interesting finding is that when the dynamics are too stiff or \(\Delta t\) is too large, lower-order integrators can quickly become unstable, while state-prediction models can maintain an inaccurate state over time. As as whole, we find that neural surrogates can be faster than un-optimized numerical solvers, but no neural surrogate is currently faster or more accurate than a highly-optimized numerical solver. While this is rather discouraging for neural surrogates, it does reflect similar findings from other works \cite{dresdner2023learningcorrectspectralmethods, McGreivy_2024} and future work in adaptive step-sizing or novel architectures can help to further accelerate neural surrogates in this regime. 

\subsection{Limitations}
As a final remark, we note some limitations of derivative prediction. Not all time-stepping schemes can be implemented with derivative prediction; in particular, it is not obvious how to extend this work to more sophisticated schemes such as implicit integrators. Furthermore, for steady-state prediction problems or boundary value problems, which could encompass phenomena such as Darcy flow or linear elasticity, derivative prediction would not work in its current formulation. Lastly, derivative prediction introduces temporal discretization constraints into model inference. This contradicts one of the main benefits of neural surrogates, which is that they accelerate simulation by taking large timesteps at the cost of convergence or accuracy guarantees. When considering this full picture, state and derivative prediction have their unique benefits and drawbacks, and the most suitable framework will depend heavily on the given problem and desired behavior of the neural surrogate. We hope that the experiments and discussion can help practitioners understand when and how to apply derivative and state prediction. 

\begin{table}[t!]
    \centering
    \begin{tabular}{l l c c c c | c c c c | c c c}
    \toprule
          & Model: & \multicolumn{4}{c}{FNO} & \multicolumn{4}{c}{Unet} & \multicolumn{3}{c}{Solver}\\
          \cmidrule(lr){3-6} \cmidrule(lr){7-10} \cmidrule(lr){11-13}
         Experiment & Metric & State & Adams & Heun & RK4 &  State & Adams & Heun & RK4 & Full & Half & Quarter\\
         \midrule
         \multirow{ 2}{*}{NS} & Runtime (s) & 0.163 & 0.158 & 0.301 & 0.622 & 0.359 & 0.359 & 0.691 & 1.379 & 3.508 & 2.692 & 2.549 \\ 
         & Rollout Error & 0.715 & 0.100 & 0.100 & 0.100 & 0.099 & 0.048 & 0.049 & 0.049 & 0.000 & 0.096 & 0.285 \\ 
          \midrule
         \multirow{ 2}{*}{Kolm. Flow} & Runtime (s) & 0.152 & 0.155 & 0.308 & 0.616 &  0.204  & 0.201 & 0.411 & 0.818 & 0.116 & - & - \\ 
         & Rollout Error & 0.707 & 1e9 & 0.452 & 0.449 & 0.810 & 1e8 & 90.05 & 0.353 & 0.000 & - & - \\ 
         \bottomrule
    \end{tabular}
    \caption{\textbf{Computational Cost.} Comparison of computational cost of different models and a baseline solver on the Navier-Stokes and Kolmogorov Flow experiments. Runtimes are reported in seconds (s) for a full rollout, averaged for each sample in the validation set. Rollout errors are given as relative L2 error, with some models being unstable. Furthermore, solver baselines are coarsened to match neural surrogate accuracies when needed by solving at full (\(n_x\times y\)), half (\(\frac{n_x}{2}\times \frac{n_y}{2}\)), and quarter (\(\frac{n_x}{4}\times \frac{n_y}{4}\)) spatial resolutions.}
    \label{tab:timing}
\end{table}

\section{Conclusion \label{sec:conclusion}} 
We evaluate framing neural surrogates to predict temporal derivatives and use an ODE integrator during inference to produce solutions. Although this idea has been studied informally, this is the first work to broadly apply this framework and investigate its benefits and limitations. We find that derivative prediction can improve model accuracy and stability across architectures and PDE problems, and has useful data and inference flexibility. Furthermore, we compare this framework with other training modifications and study the effect of temporal resolution on model performance. Lastly, we hypothesize potential explanations for the improved stability and accuracy of derivative prediction and study the sources of error in derivative prediction. We hope that future work can apply derivative prediction across a wider variety problems and model architectures, and improve performance at coarser step sizes.

\section{Declarations}
\bmhead{Data availability}
Datasets used in this work are provided at: \url{https://huggingface.co/datasets/ayz2/temporal_pdes}

\bmhead{Code availability}
Code to reproduce this work is provided at: \url{https://github.com/anthonyzhou-1/temporal_pdes}

\bmhead{Funding}
This research did not receive any specific grant from funding agencies in the public, commercial, or not-for-profit sectors.


\newpage 

\bibliography{main}

\end{document}